\documentclass[preprint,12pt]{elsarticle}

\usepackage{amssymb}
\usepackage{amsmath}
\usepackage{multirow} %表格
\usepackage{booktabs}
\usepackage[normalem]{ulem} %下划线
\usepackage[dvipsnames]{xcolor}
\definecolor{mygray}{gray}{0.6}
\usepackage{subfigure}
\usepackage{bbm} %指示函数包

\usepackage{amssymb}
\usepackage{mathtools}
\usepackage{amsthm}
\usepackage[ruled,linesnumbered]{algorithm2e}
\usepackage{hyperref}%超链接

\usepackage[capitalize,noabbrev]{cleveref}
\usepackage{soul}
\usepackage{lineno}  %行号

\journal{}

\begin{document}

%\begin{titlepage}
%    \centering
%    \vspace*{2.5cm}  % 上方留白略小，整体更居中
%
%    % 标题部分
%    {\LARGE CCFC: Bridging Federated Clustering and Contrastive Learning \par}
%
%    \vspace{2.8cm}  % 标题与作者之间空白更协调
%
%    % 作者信息
%    {\normalsize
%        Jing Liu (jingl@email.cufe.edu.cn)\\[0.7em]
%        Jie Yan (jieyan@email.cufe.edu.cn)\\[0.7em]
%        Zhong-Yuan Zhang (zhyuanzh@cufe.edu.cn)%\\[1ex]
%    }
%
%    \vspace{1cm}  % 作者与单位之间空白适中
%
%    % 单位信息
%    {\normalsize
%        School of Statistics and Mathematics,\\[0.7em]
%        Central University of Finance and Economics,
%        Beijing, P.R. China
%    }
%
%    \vfill
%\end{titlepage}

\begin{frontmatter}

%% Title, authors and addresses

%% use the tnoteref command within \title for footnotes;
%% use the tnotetext command for theassociated footnote;
%% use the fnref command within \author or \affiliation for footnotes;
%% use the fntext command for theassociated footnote;
%% use the corref command within \author for corresponding author footnotes;
%% use the cortext command for theassociated footnote;
%% use the ead command for the email address,
%% and the form \ead[url] for the home page:
%% \title{Title\tnoteref{label1}}
%% \tnotetext[label1]{}
%% \author{Name\corref{cor1}\fnref{label2}}
%% \ead{email address}
%% \ead[url]{home page}
%% \fntext[label2]{}
%% \cortext[cor1]{}
%% \affiliation{organization={},
%%             addressline={},
%%             city={},
%%             postcode={},
%%             state={},
%%             country={}}
%% \fntext[label3]{}

\title{CCFC: Bridging Federated Clustering and Contrastive Learning}

%% use optional labels to link authors explicitly to addresses:
%% \author[label1,label2]{}
%% \affiliation[label1]{organization={},
%%             addressline={},
%%             city={},
%%             postcode={},
%%             state={},
%%             country={}}
%%
%% \affiliation[label2]{organization={},
%%             addressline={},
%%             city={},
%%             postcode={},
%%             state={},
%%             country={}}
\begin{titlepage}

\author[label1]{Jing Liu}
\ead{jingl@email.cufe.edu.cn}

\author[label1]{Jie Yan}
\ead{jieyan@email.cufe.edu.cn}

\author[label1]{Zhong-Yuan Zhang\corref{mycorrespondingauthor}}
\ead{zhyuanzh@cufe.edu.cn}

\cortext[mycorrespondingauthor]{Corresponding author.}

\address[label1]{School of Statistics and Mathematics, \\ Central University of Finance and Economics, Beijing, P.R.China}
\end{titlepage}

%\author{123} %% Author name
%
%%% Author affiliation
%\affiliation{organization={},%Department and Organization
%            addressline={},
%            city={},
%            postcode={},
%            state={},
%            country={}}

%% Abstract
\begin{abstract}
%% Text of abstract
Federated clustering, an essential extension of centralized clustering for federated scenarios, enables multiple data-holding clients to collaboratively group data while keeping their data locally. In centralized scenarios, clustering driven by representation learning has made significant advancements in handling high-dimensional complex data. However, the combination of federated clustering and representation learning remains underexplored. To bridge this, we first tailor a cluster-contrastive model for learning clustering-friendly representations. Then, we harness this model as the foundation for proposing a new federated clustering method, named cluster-contrastive federated clustering (CCFC). Benefiting from representation learning, the clustering performance of CCFC even \textit{double} those of the best baseline methods in some cases. Compared to the most related baseline, the benefit results in substantial NMI score improvements of up to 0.4155 on the most conspicuous case. Moreover, CCFC also shows superior performance in handling device failures from a practical viewpoint. Code is available at \href{https://github.com/Jarvisyan/CCFC-pytorch}{https://github.com/Jarvisyan/CCFC-pytorch}.
\end{abstract}

%%%Graphical abstract
%\begin{graphicalabstract}
%%\includegraphics{grabs}
%\end{graphicalabstract}
%
%%%Research highlights
%\begin{highlights}
%\item Research highlight 1
%\item Research highlight 2
%\end{highlights}

%% Keywords
\begin{keyword}
%% keywords here, in the form: keyword \sep keyword
Federated clustering, representation learning, contrastive learning.
%% PACS codes here, in the form: \PACS code \sep code

%% MSC codes here, in the form: \MSC code \sep code
%% or \MSC[2008] code \sep code (2000 is the default)

\end{keyword}

\end{frontmatter}

%% Add \usepackage{lineno} before \begin{document} and uncomment
%% following line to enable line numbers
%% \linenumbers

%% main text
%%

%\linenumbers
\section{Introduction}
Federated clustering (FC), an essential extension of centralized clustering for federated scenarios, enables multiple data-holding clients to collaboratively group data while keeping their data locally \cite{dennis2021heterogeneity, stallmann2022towards, sdafc, ppfcgan}. It often constitutes a significant initial step in many learning tasks, such as client-selection \cite{fu2023client} and personalization \cite{long2023multi, cho2023communication}.

As an extension of centralized clustering, it is natural to extend centralized clustering methodologies into FC. For instance, k-fed \cite{dennis2021heterogeneity} is an extension of k-means (KM) \cite{lloyd1982least}, while federated fuzzy c-means (FFCM) \cite{stallmann2022towards} builds upon fuzzy c-means (FCM) \cite{bezdek1984fcm}. Both of these extended approaches employ a similar methodology: alternately estimating local and global cluster centroids, and labeling the local data based on their proximity to these global centroids. Nevertheless, the global cluster centroids are vulnerable to heterogeneous client data, diminishing both the robustness and overall performance of the model \cite{sdafc}. To tackle this, synthetic data aided federated clustering (SDA-FC) \cite{sdafc} and privacy-preserving federated deep clustering based on GAN (PPFC-GAN) \cite{ppfcgan} bridged FC and deep generative model, enabling the application of centralized clustering techniques to synthetic data, and safeguarding global cluster centroids from data heterogeneity problem. Although these methods have advanced FC, a substantial gap persists between FC and centralized clustering. For instance, on the MNIST dataset, the NMI \cite{strehl2002cluster} values of the top 10 centralized clustering methods surpass 0.93\footnote{https://paperswithcode.com/sota/image-clustering-on-mnist-full}. In stark contrast, even in the simplest federated scenario, where local data among clients is independently-and-identically-distributed (IID), the NMI value of the cutting-edge PPFC-GAN falls short of 0.7 (\cref{NMI}).

%\begin{figure}[!t]
%\centering
%%\vspace{-3cm}
%\subfigure[Raw pixels]{
%\includegraphics[height = 3.5cm, width = 3.9cm]{figures/raw_tsne_local.png}
%\label{case1a}}
%\quad
%\subfigure[PPFC-GAN]{
%\includegraphics[height = 3.5cm, width = 4cm]{figures/latent_sda_fdc_tsne_local.png}
%\label{case1b}}
%
%\subfigure[Standalone SimSiam]{
%\includegraphics[height = 3.5cm, width = 4cm]{figures/latent_p0_scfc_client0_tsne.png}
%\label{case1c}}
%\,
%\subfigure[CCFC]{
%\includegraphics[height = 3.5cm, width = 4cm]{figures/latent_ccfc_trained_tsne_local.png}
%\label{case1d}}
%
%\caption{\ul{\textbf{t-SNE visualizations on MNIST (best viewed in color)}. \textbf{(a)} shows the local data in the original data space, where each color corresponds to a specific category of digits in MNIST. \textbf{(b) - (c)} show the local data in different latent spaces.}}
%\label{case1}
%\end{figure}

\begin{figure}
\centering
\includegraphics[height = 7cm, width = 10cm]{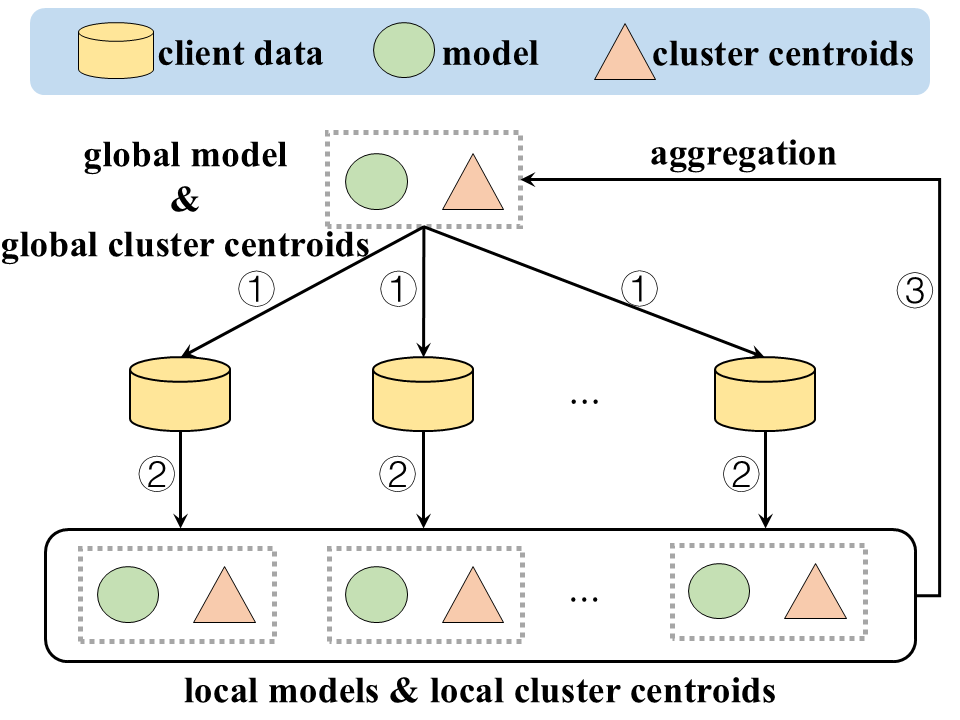}
\caption{\textbf{CCFC architecture}. The numbers indicate the order of the corresponding steps in each communication round.}
\label{ccfc}
\end{figure}

To bridge this gap, it is logical to revisit centralized clustering methodologies, especially the most advanced ones. In centralized scenarios, clustering driven by representation learning has made significant advancements in handling high-dimensional complex data,  particularly images \cite{zhou2022comprehensive, ren2024deep}. However, the marriage of federated clustering and representation learning (e.g. contrastive learning \cite{oord2018representation, bachman2019learning}) remains an underexplored research avenue. As PPFC-GAN has validated the feasibility of training deep learning model using shared synthetic data, a obvious next step is to explore contrastive learning on this data. However, the increased utility of synthetic data often comes with a heightened risk of deep data leakage from shared synthetic data \cite{shao2023survey}. In contrast to synthetic data, which may serve as replicas of real data and thus pose a direct privacy risk \cite{ghalebikesabi2023differentially}, sharing model parameters offers a significantly greater degree of privacy preservation. Recall that the core distinction between FC and centralized clustering lies in privacy preservation.  Hence, we choose to use the classical FedAvg framework \cite{mcmahan2017communication} to bridge FC and
contrastive learning, rather than synthetic data aided framework \cite{sdafc, ppfcgan}.

Motivated by these, we first tailor a contrastive model, named cluster-contrastive model, for learning clustering-friendly representations that show improved intra-cluster compactness and inter-cluster separability. Then, we harness this model as the foundation for proposing a simple but effective federated clustering method, named \textbf{cluster-contrastive federated clustering (CCFC)}, which embeds the cluster-contrastive model into the FedAvg framework. Throughout the entire training procedure of CCFC, the only shared information between clients and the server is cluster-contrastive models and cluster centroids, thereby preserving data privacy. As shown in \cref{ccfc}, CCFC involves three main steps in each communication round: global information dissemination, local training and local information aggregation. In the first step, each client first downloads a global model and $k$ global cluster centroids from the server, and then update the local model with the downloaded one, where $k$ is the number of clusters. In the second step, each client performs cluster assignment by labeling their local data with the index of the nearest global cluster centroid in the latent space, and they train the local model using both the local data and the assignment results. After this, KM is employed to create $k$ local cluster centroids for storing local semantic information. In the final step, clients upload their trained local models and local cluster centroids to the server, which aggregates these local models into a new global model through weighted averaging, and aggregates these local cluster centroids into $k$ new global cluster centroids through KM.

As widely acknowledged, the performance of the FedAvg framework can be hindered by the issue of data heterogeneity, and this problem remains unresolved and challenging \cite{shao2023survey, ma2022state, lu2024federated}. Thus, this work focuses on how contrastive learning aids in FC and attempts to analyze how data heterogeneity impairs the clustering-friendliness of the learned representations, while addressing this issue of data heterogeneity is deferred to future work. Comprehensive experiments demonstrate the superiority of the cluster-contrastive model in learning clustering-friendly representations, as well as the excellence of the proposed clustering method in terms of clustering performance and handling device failures. The most related method to our CCFC is k-fed \cite{dennis2021heterogeneity}, with the main distinction being the inclusion of a representation learning process in CCFC. Surprisingly, this learning process results in substantial NMI \cite{strehl2002cluster} score improvements of up to 0.4155 and Kappa \cite{liu2019evaluation} score enhancements of 0.4593 on the most conspicuous case. With the rise of data heterogeneity, the learned representations progressively lose their clustering-friendly traits, as evidenced by \textit{semantic inconsistency} in the latent spaces of local models and overlapping cluster representations (\textit{cluster collapse}). %Resolving these two unfavorable phenomena may further unleash the potential of CCFC, which is left for future work.}

In summary, our contributions are as follows:
\begin{itemize}
\item
A tailored cluster-contrastive model is introduced to facilitate the acquisition of more clustering-friendly representations.

\item
Leveraging this model, a simple yet effective federated clustering method, CCFC, is proposed.

\item
Comprehensive experiments validate the significant superiority of CCFC and offer valuable insights into its performance.

\item
A systematic analysis of how data heterogeneity impairs the clustering-friendliness of the learned representations offers a potential direction for alleviating the data heterogeneity problem.
\end{itemize}

The remainder of this paper is structured as follows: \cref{sect2} provides a concise review of representative federated clustering methods and discusses the current state of contrastive learning research in federated scenarios. \cref{sect3} presents a new cluster-contrastive model, upon which the CCFC, a simple yet effective federated clustering method, is developed. \cref{sect4} highlights the advantages of the proposed methods and conducts a comprehensive analysis of how data heterogeneity impairs the clustering-friendliness of the learned representations. Finally, \cref{sect5} concludes the paper.  %\mbxof{\ref{sect2}}

\section{Related Work}
\label{sect2}
In this section, we first provide a concise review of representative federated clustering methods and then discuss the current state of contrastive learning research in federated scenarios.

\subsection{Clustering}
Clustering, a fundamental machine learning task, aims to group similar samples together and often constitutes a significant initial step in many centralized or federated learning tasks, including information retrieval \cite{zhao2022centerclip, anju2022faster}, anomaly detection \cite{jain2022k, javaheri2023fuzzy}, client-selection \cite{fu2023client} and personalization \cite{long2023multi, cho2023communication}. Traditionally, it is assumed that data is consolidated on a central server for model training. However, in federated scenarios, data is distributed among multiple isolated clients, and the sharing of local data is prohibited due to privacy concerns. In this scenario, clustering tasks are fraught with significant difficulties, as local data alone is inadequate for the accurate clustering of itself, and a centralized global dataset remains unfeasible \cite{stallmann2022towards}.

To address these, federated clustering (FC) has arisen, allowing multiple clients to collaboratively group data while preserving their data locally \cite{dennis2021heterogeneity, stallmann2022towards, sdafc, ppfcgan}. As an extension of centralized clustering, FC inherently embodies an exploratory path, i.e. extending centralized clustering methodologies to federated scenarios. Representative extensions include k-FED \cite{dennis2021heterogeneity} and SDA-FC-KM \cite{sdafc}, derived from k-means (KM) \cite{lloyd1982least}; FFCM \cite{stallmann2022towards} and SDA-FC-FCM \cite{sdafc}, extensions of fuzzy c-means (FCM) \cite{bezdek1984fcm}; and PPFC-GAN \cite{ppfcgan}, an outgrowth of DCN \cite{yang2017towards}. The key idea behind these extensions is similar: constructing $k$ global cluster centroids and labeling local data with the index of the nearest global centroid, where $k$ is the number of clusters. In methods k-FED and FFCM, each client first employs KM or FCM to extract local cluster centroids from their local data and then transmit these centroids to the server, where these local centroids are merged into $k$ global centroids through KM. Nonetheless, the global cluster centroids are fragile and sensitive to heterogeneous data among clients, which undermines the model's robustness and model performance. To address this, SDA-FC-KM, SDA-FC-FCM, and PPFC-GAN bridged FC and deep generative model, enabling the application of centralized clustering techniques to synthetic data for the construction of $k$ global cluster centroids. Although these extensions have advanced FC, a substantial gap persists between FC and centralized clustering. For instance, on the MNIST dataset, the NMI values of the top 10 centralized clustering methods surpass 0.93\footnote{https://paperswithcode.com/sota/image-clustering-on-mnist-full}. In stark contrast, even in the simplest federated scenario, where local data among clients is independently-and-identically-distributed (IID), the NMI value of the cutting-edge PPFC-GAN falls short of 0.7 (\cref{NMI}).

%Although the synthetic data can make the model immune to the data heterogeneity problem without sharing private data,

%在中心化场景，IID假设。为了简化分析，一个常见的做法就是假设IID, 但即使是在这么简单的情景中也不行。。

%Specifically, k-fed and FFCM employ KM or FCM to mine local cluster centroids on clients, and utilize KM on the server to integrate these local centroids into $k$ global centroids, where $k$ is the number of clusters. Then, cluster assignments can be made by assigning local data points to the nearest global centroid. However, the integrated global centroids are sensitive to data heterogeneity among clients, leading to non-robust and suboptimal clustering performance \cite{sdafc}.

In the realm of centralized clustering, the rapid progress can largely be attributed to the incorporation of representation learning techniques \cite{zhou2022comprehensive}. Hence, it stands to reason that federated clustering methods powered by representation learning may hold the key to narrowing this existing gap. However, the combination of federated clustering and representation learning remains underexplored.

\subsection{Contrastive Learning}
%why表示学习 -> 关键是代理任务 -> 我们设计了一个新的代理任务
Contrastive learning \cite{oord2018representation, bachman2019learning} has emerged as a prominent paradigm within the realm of representation learning, particularly for the acquisition of visual representations. In early studies \cite{chen2020simple, he2020momentum}, these methods aims to maximize the similarity between augmented views of the same image (i.e., \textit{positive pairs}) while minimizing the similarity between augmented views of different images (i.e., \textit{negative pairs}). In these methods, the contrast between the negative pairs is considered a key factor in preventing model collapse (i.e., the model possesses a trivial solution where all outputs collapse to a constant). Nevertheless, further studies \cite{grill2020bootstrap, chen2021exploring} have shown that the negative pairs are not essential for averting model collapse. While contrastive learning has been extensively explored and proven effective in centralized scenarios \cite{ericsson2022self, schiappa2023self, mohamed2022self}, its extension into federated scenarios has been relatively limited. These extensions can be broadly categorized into two groups: one employs contrastive learning to handle the challenges posed by the non-IID problem in federated classification tasks \cite{li2021model, tan2022fedproto, mu2023fedproc}, while the other harnesses it for the acquisition of generic representations for downstream learning tasks \cite{zhuang2021collaborative, han2022fedx, zhuang2022divergence}.

In FC, there is a concurrent work called federated momentum contrastive clustering (FedMCC) \cite{miao2024federated}, which is an extension of contrastive clustering (CC) \cite{li2021contrastive}. CC is a centralized clustering approach that explicitly executes contrastive learning at both instance and cluster levels by maximizing positive pair similarities and minimizing negative pair similarities in the row and column representation spaces, respectively. Note that the objective function of CC, besides contrastive loss terms, incorporates an entropy-based regularization term to prevent trivial solutions and encourage uniform cluster sizes \cite{li2021contrastive, hu2017learning}. This technique to avoid trivial solutions has been carried over in FedMCC. Although many real-world datasets in centralized scenarios, like Fashion-MNIST, CIFAR-10 \cite{krizhevsky2009learning}, and STL-10 \cite{coates2011analysis}, feature a uniform class distribution, the training data distribution in federated environments is generally imbalanced and heterogeneous due to variations in client preferences \cite{wang2021addressing, hamidi2024fed}. Consequently, encouraging uniform cluster sizes in FedMCC becomes unreasonable, thus restricting its real-world applicability. %\ul{Moreover, FedMCC is an CC extension based on the classical Federated Averaging (FedAvg) \mbox{\cite{mcmahan2017communication}} framework, where individual clients train the model locally, and the server aggregates updates globally. Iterative client-server communication will continue until the model converges. Nevertheless, data heterogeneity may diminish the performance of the updated local model in the $t + 1$ communication round compared to the aggregated (global) model in the $t$ round, leading to model regression \mbox{\cite{li2021model}}. Our empirical findings reveal that the model regression also occurs in the IID scenario (refer to \mbox{\cref{case1}}). }
%In terms of data augmentation, FEDFA+ shares similarities with a concurrent work [21] that also employs low-order statistics for augmentation in FL. However, FEDFA+ distinguishes itself by utilizing a probabilistic formulation and demonstrating higher communication efficiency.

By contrast, our study introduces a new contrastive model for clustering tasks, called cluster-contrastive model, which places no limitations on cluster sizes, thus offering greater flexibility. Leveraging this model, we further propose a simple yet effective federated deep clustering method named cluster-contrastive federated clustering (CCFC). %\ul{As for the model regression problem, we incorporate a model-contrastive regularization term to constrain the local updates, ensuring closer alignment with the global model aggregated in the last communication round (see \mbox{\cref{case4}} for effectiveness).}

\section{Cluster-Contrastive Federated Clustering}
\label{sect3}
In this section, we first propose a cluster-contrastive model to encode the semantic structures discovered by clustering into the latent representation space. Then, we harness this model as the foundation for proposing a simple but effective federated clustering method.

\subsection{Cluster-Contrastive Model}
In this work, due to the elegance of the SimSiam \cite{chen2021exploring}, we select it as the prototype model among numerous contrastive models. It maximizes the similarity between augmented views of the same image (i.e., \textit{positive pairs}) directly, without minimizing the similarity between augmented views of different images (i.e., \textit{negative pairs}) or using a momentum encoder. Contrasting negative pairs usually requires some customized techniques (such as memory bank \cite{wu2018unsupervised} and memory queue \cite{he2020momentum}) to perform large-batch training.  The momentum encoder \cite{grill2020bootstrap} introduces extra hyperparameter and necessitates additional memory allocation for storing historical momentum information. Both of these can complicate the model training and incur extra computational costs. However, the representations learned by SimSiam may be suboptimal for clustering tasks, as it leans towards grasping low-level semantics through sample-level differences among images, neglecting the cluster-level distinctions among them.

%the proposed model leans towards grasping low-level semantics through the utilization of sample-level disparities among images, and the resulting representations are suboptimal for clustering tasks.  In contrast, the latent representation learned by the generalized SimSiam (cluster-contrastive model), which takes into account the cluster-level distinctions among images, is more clustering-friendly.

%the similarity between augmented views of the same image (i.e., \textit{positive pairs}) while minimizing the similarity between augmented views of different images (i.e., \textit{negative pairs})

%It does not require the negative pairs like MOCO \cite{he2020momentum} or SimCLR \cite{chen2020simple}, nor does it necessitate the use of a momentum encoder like BYOL \cite{grill2020bootstrap}.

\begin{figure}
\centering
\includegraphics[height = 5cm, width = 9cm]{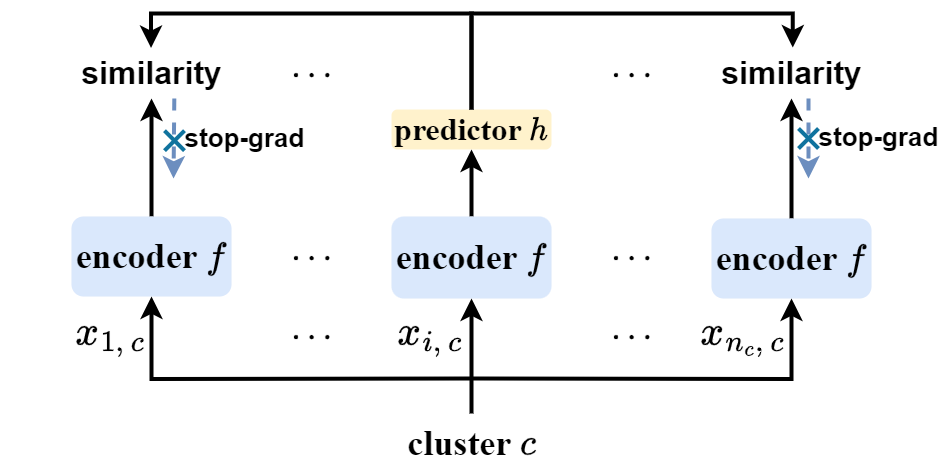}
\caption{\textbf{Cluster-contrastive model architecture}. $n_c$ images from cluster $c$ are encoded by the same encoder $f$, which comprises a backbone (e.g. ResNet-18) and an MLP. Then, an MLP predictor is applied to the $i$-th path, while $n_c - 1$ stop-gradient operations are applied to the remaining paths. The model maximizes the average similarity of $x_{i, c}$ to other images.}
\label{cc_arch}
\end{figure}

%负样本对的坏处

To handle this, we tailor a new contrastive model for clustering tasks by extending SimSiam, called \textbf{cluster-contrastive model}. It maximizes the similarity among images within the same cluster to  learn cluster-invariant representations, i.e. samples within the same cluster are expected to have similar latent representations. Formally, given $k$ clusters obtained from clustering, each one comprises $n_c$ images. As shown in \cref{cc_arch}, the proposed model takes $n_c$ images from cluster $c$ as inputs. These images are processed by the same encoder $f$, which comprises a backbone (e.g., ResNet-18 \cite{he2016deep}) and an MLP projector \cite{chen2020simple}. The latent representations of these images are denoted as $z_{i,\, c} = f(x_{i,\, c}),\, i = 1,\, \cdots,\, n_c$. Then, an MLP predictor \cite{grill2020bootstrap}, denoted as $h$, transforms the $i$-th latent representation and matches it to the remaining latent representations. The prediction of $x_i$ for the remaining images is denoted as $p_{i,\, c} = h(f(x_{i,\, c}))$. The negative average cosine similarity of
$x_{i,\, c}$ to other images is defined as:
\begin{equation}
D(x_{i,\, c}, \{x_{j,\, c}\}_{j \neq i}^{n_c}) = - \frac{1}{n_c - 1}  \sum_{j \neq i}^{n_c}  \frac{p_{i,\, c}}{\|p_{i,\, c}\|_2}  \cdot  \frac{z_{j,\, c}}{\|z_{j,\, c}\|_2},
\end{equation}
where $\|\cdot\|_2$ is $\ell_2$-norm. Finally, the overall loss is defined as:
\begin{equation}
\mathcal{L} = \sum_{c = 1}^{k} \frac{1}{kn_c} \sum_{i = 1}^{n_c} D\left(x_{i,\, c},\, stopgrad\left(\{x_{j,\, c}\}_{j \neq i}^{n_c}\right)\right),
\label{my_sim}
\end{equation}
where the stop-gradient operation ($stopgrad(\cdot)$) is an critical component to avoid model collapse when the contrastive model is trained exclusively with the positive pairs \cite{chen2021exploring}. Specifically, $stopgrad(\{x_{j,\, c}\}_{j \neq i}^{n_c})$ entails treating each element in $\{z_{j,\, c}\}_{j \neq i}^{n_c}$ as a constant, ensuring no gradient flows from $z_{j,\, c}$ to $x_{j,\, c}$ in this term.

%Actually, the proposed model is a more general form of simple Siamese (SimSiam) networks \cite{chen2021exploring}. Consider a scenario where $k = n$ ($n$ is the number of images) and $n_c = 2$, signifying that each image forms a cluster, with each cluster consists of the two augmented views of that image, the proposed model reverts to SimSiam. In this context, the proposed model leans towards grasping low-level semantics through the utilization of sample-level disparities among images, and the resulting representations are suboptimal for clustering tasks.  In contrast, the latent representation learned by the generalized SimSiam (cluster-contrastive model), which takes into account the cluster-level distinctions among images, is more clustering-friendly.

\subsection{Cluster-Contrastive Federated Clustering}
%the key is ...构造质心
%As the most straightforward approach to integrating FC with contrastive learning involves each client independently training a local model on their local data, we will first demonstrate the necessity and challenge of employing FedAvg for contrastive learning before delving into CCFC.

\begin{figure}[!t]
\centering
%\vspace{-3cm}
\subfigure[Raw pixels]{
\includegraphics[height = 3.3cm, width = 6cm]{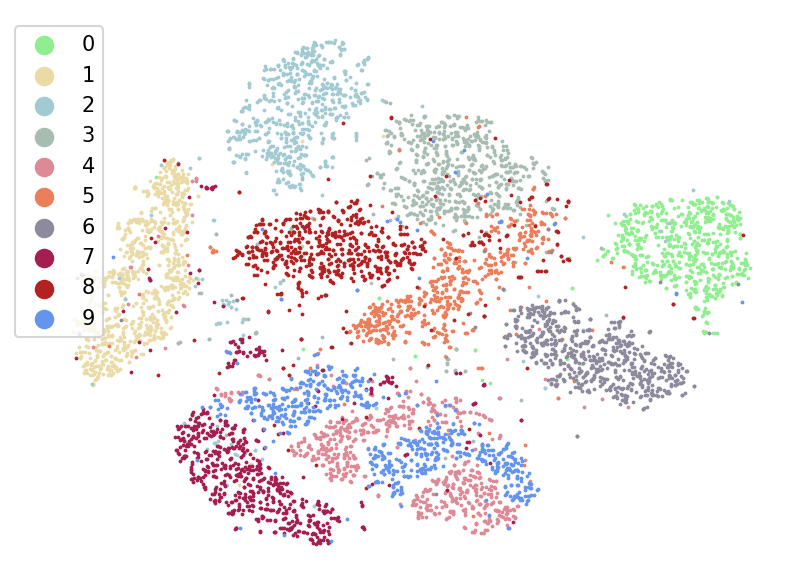}
\label{case1a}}
\quad
\subfigure[Standalone local model]{
\includegraphics[height = 3.3cm, width = 6cm]{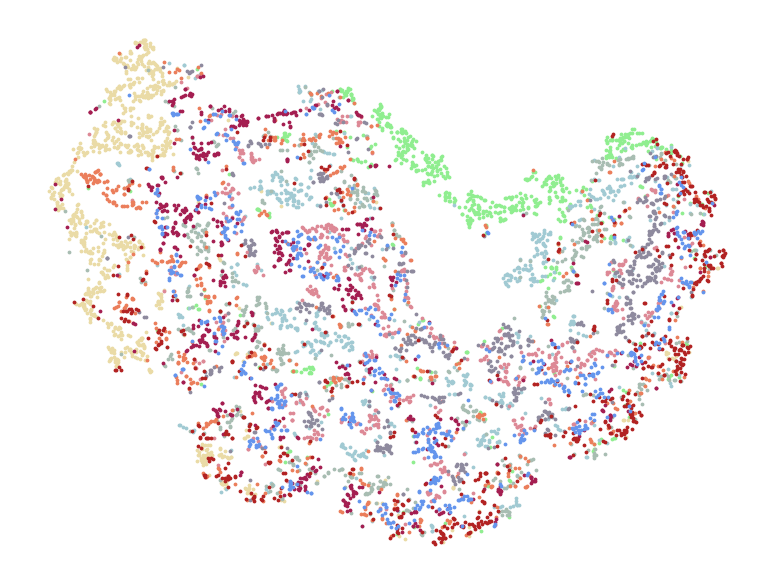}
\label{case1b}}

\subfigure[FedAvg global model]{
\includegraphics[height = 3.3cm, width = 6cm]{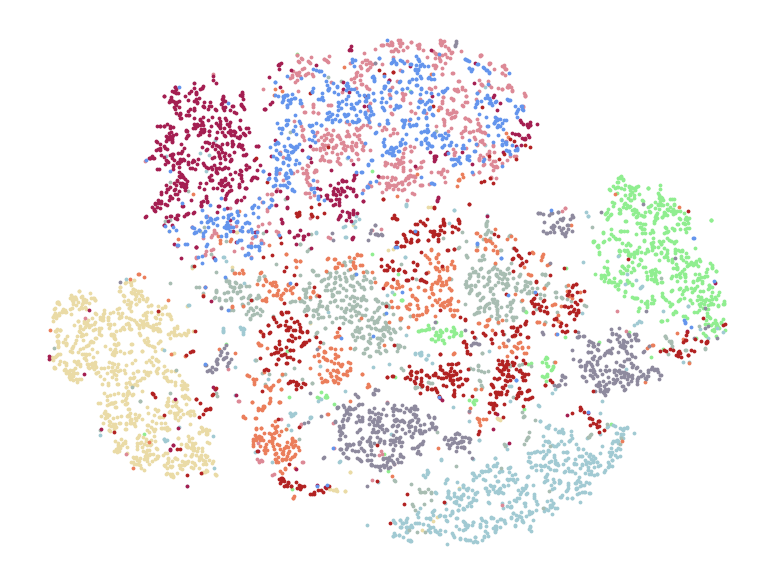}
\label{case1c}}
\,
\subfigure[FedAvg local model]{
\includegraphics[height = 3.3cm, width = 6cm]{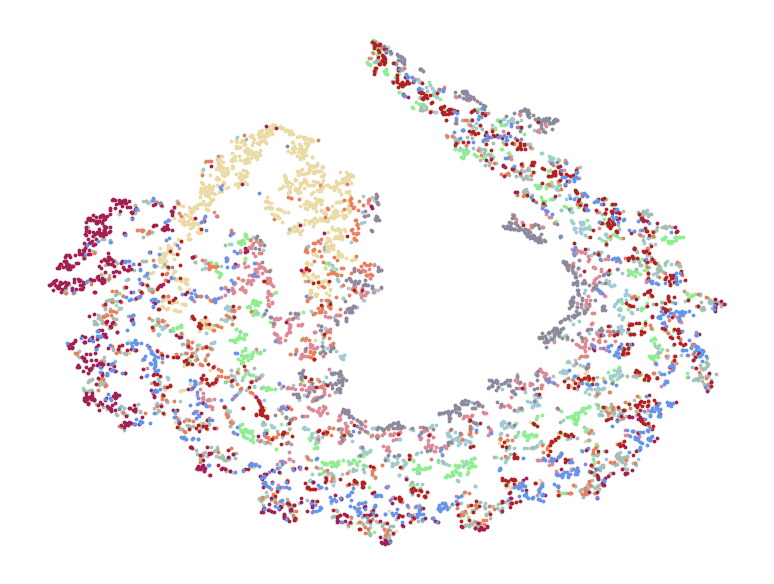}
\label{case1d}}

\caption{\textbf{t-SNE visualizations on MNIST (best viewed in color)}. \textbf{(a)} shows the local data in the original data space, where each color corresponds to a specific category of digits in MNIST. \textbf{(b) - (c)} show the local data in different latent spaces.}
\label{case1}
\end{figure}

Given a real-world dataset $X$ distributed among $m$ clients, i.e., $X=\bigcup_{l=1}^{m} X^l$. There are two straightforward methods for integrating FC with contrastive learning here: 1) Independent execution of contrastive learning by each client; 2) Integration of contrastive learning into the Federated Averaging (FedAvg) \cite{mcmahan2017communication} framework. To illustrate the necessity of the integration and the challenges it poses, we simulate a simple federated scenario on MNIST, where images are evenly distributed among 10 clients. \cref{case1a} showcases the t-SNE visualization of a randomly selected local dataset in its original data space, and one can observe that many samples from different categories are mixed up together, necessitating representation learning. As shown in \cref{case1b} and \cref{case1c}, one can see that: 1) The local model trained solely on local data exhibits poor performance. 2) The collaborative training yields a superior global model. However, in the subsequent round of local updates, the limited information from the local data can lead to model regression (\cref{case1d}). A similar problem of the model regression has also been observed in federated classification tasks, and it can be alleviated by introducing a model-contrastive loss to regularize the local model to refrain from deviating excessively from the global model during the local training \cite{li2021model}.

Motivated by these, we propose a new federated clustering model, named as \textbf{cluster-contrastive federated clustering (CCFC)}, which embeds the cluster-contrastive model into the FedAvg framework with the model-contrastive regularization term. As shown in \cref{ccfc} and \cref{ps_code}, CCFC involves three steps in each communication round: global information dissemination, local training and local information aggregation. Details are given below.

\noindent \textbf{Global information dissemination.} Each client $l$ ($l = 1,\, 2,\ \cdots, \, m$) downloads a global model $w^g = (f^g,\, h^g)$, and $k$ global cluster centroids $\{\eta_c^g\}_{c = 1}^{k}$ from the server. Here, $f^g$ is the global encoder network and $h^g$ is the global predictor network. The local model $w^l = (f^l,\, h^l)$ is updated with the global model $w^g$.

\noindent \textbf{Local training.} To acquire clustering-friendly representations using the cluster-contrastive model, each client $l$ first performs cluster assignment by labeling their local data with the index of the nearest global cluster centroid in the latent space, as defined in:
\begin{equation}
\underset{c=\{1,\, \cdots,\, k\}}{\arg \min }\left\|f^{g}(x) - \eta^{g}_{c}\right\|_{2}.
\label{label_assign}
\end{equation}

Then, each client $l$ trains the local model $w^l$ using their local data $X^l$ and the assignment results. To alleviate the model regression problem, we introduce a model-contrastive regularization term into \cref{my_sim}. Finally, the revised loss function is denoted as:
\begin{equation}
\mathcal{L_{R}} = \mathcal{L} + \sum_{c = 1}^{k} \frac{\lambda}{kn_c} \sum_{i = 1}^{n_c} R(w^l(x_{i,\, c}),\, w^g(x_{i,\, c})) ,
\label{my_sim_revised}
\end{equation}
where
\begin{align}
R(w^l(x_{i,\, c}),\, w^g(x_{i,\, c})) & =  - \frac{p^l_{i,\, c}}{\|p^l_{i,\, c}\|_2}  \cdot  \frac{p^g_{i,\, c}}{\|p^g_{i,\, c}\|_2} \\
p^l_{i,\, c} & = h^l(f^l(x_{i,\, c})), \\
p^g_{i,\, c} & = h^g(f^g(x_{i,\, c})),
\end{align}
and $\lambda$ is the tradeoff hyperparameter. By minimizing \cref{my_sim_revised}, the first term will encourage the local model to encode the local semantic structures discovered by the assignment result into the latent representation space. Note that the parameters of the global model $w^g$ remain fixed throughout the local training, whereas only those of the local model $w^l$ undergo updates. Consequently, the second term will correct the local updates by maximizing the agreement (cosine similarity) of representation learned by the current local model $w^l$ and that learned by the global model $w^g$ from the previous communication round.

After training, for the subsequent integration of local semantic information, each client first employs their respective local model to encode the local data into the latent representation space, and then use k-means (KM) to create $k$ local cluster centroids for storing the local semantic information.

\noindent \textbf{Local information aggregation.} First, each client uploads their trained local model and local cluster centroids to the server. Then, the server aggregates these local models into a new global model by computing their weighted average, as follows:
\begin{equation}
w^g = \sum_{l = 1}^{m} \frac{|X^l|}{|X|}w^l.
\label{agg_m}
\end{equation}
And $k$ new global cluster centroids can be obtained by applying KM to these local cluster centroids.

So far, the whole communication loop of CCFC has been built up. And the final clustering result can be obtained by solving \cref{label_assign}.

\begin{algorithm}[!t]
\caption{CCFC}\label{ps_code}
Initialize the global model $w^g$ and the global cluster centroids $\{\eta_c^g\}_{c = 1}^{k}$. \\
\For {$round = 1,\, 2,\, \cdots,\, t$}{
\textbf{Clients execute:}\\
\For {$l = 1,\, 2,\, \cdots,\, m$ \textbf{\textup{in parallel}}}{
\textbf{Global information dissemination:}\\
\hspace{1em} download $w^g$ and $\{\eta_c^g\}_{c = 1}^{k}$ from server \\
\hspace{1em} update the local model: $w^l = w^g$\\
\textbf{Local training:}\\
\hspace{1em} group $X^l$ by solving \cref{label_assign}\\
\hspace{1em} train $w^l$ by minimizing \cref{my_sim_revised}\\
\hspace{1em} mine $k$ local cluster centroids by KM\\
\textbf{Local information aggregation:}\\
\hspace{1em} upload $w^l$ and the local cluster centroids \\
\hspace{1em} to server\\
}
\textbf{Server executes:}\\
\hspace{1.3em} \textbf{Local information aggregation:}\\
\hspace{2.3em} update $w^g$ by \cref{agg_m}\\
\hspace{2.3em} update $\{\eta_c^g\}_{c = 1}^{k}$ by  applying KM  to these\\ \hspace{2.3em} local cluster centroids.
}
\textbf{Cluster assignment}: solve \cref{label_assign}
\end{algorithm}

\section{Experiment}
\label{sect4}
In this section, we detail the experimental settings, assesses the effectiveness of our proposed methods, analyze the clustering-friendliness of the learned representations in different federated scenarios, and conduct a sensitivity analysis of clustering methods to device failures. Finally, we present a summary of the experimental findings.

\subsection{Experimental Setup}
A universally adopted benchmark dataset for federated learning remains scarce at present, because the heterogeneity of the local data distributions among clients are unknown in real-world scenarios. Following \cite{chung2022federated, sdafc, ppfcgan}, we simulate a range of federated scenarios by partitioning a real-world dataset into $k^*$ smaller subsets, each dedicated to a client, and adjusting the hyperparameter $p$ to scale the data heterogeneity for each client, where $k^*$ is the number of true clusters. Specifically, for the $l$-th client with $s$ images, $p\cdot s$ images are sampled from the $l$-th cluster, while the remaining $(1 - p) \cdot s$ images are drawn randomly from the entire data. The hyperparameter $p$ varies from 0 (data is randomly distributed among $m$ clients) to 1 (each client forms a cluster). Our experiments are conducted on four real-world datasets: MNIST (70,000 images with 10 classes), Fashion-MNIST (70,000 images with 10 classes), CIFAR-10 (60,000 images with 10 classes), and STL-10 (13,000 images with 10 classes).

Recall that the proposed cluster-contrastive model requires the training of just two network modules: the encoder (a backbone plus an MLP projector) and an MLP predictor. For MNIST and Fashion-MNIST, the encoder includes three convolutional layers, while for CIFAR-10 and STL-10, we use ResNet-18 \cite{he2016deep} as the backbone. For all the considered datasets, both the projector and predictor consist of two fully connected layers. The tradeoff hyperparameter $\lambda$ is set to 0.001, 1, 0.1 and 0.1 for MNIST, Fashion-MNIST, CIFAR-10 and STL-10, respectively. The latent representation dimensions are set to be 256, 64, 256 and 256, for the respective datasets.

The proposed method is trained using the Adam optimizer \cite{kingma2014adam}, implemented in the PyTorch \cite{paszke2019pytorch}.

\begin{table*}[!t]
\centering
%\vspace{-3cm}
%\setlength{\abovecaptionskip}{10pt}%
%\setlength{\belowcaptionskip}{10pt}%
%\captionsetup{width=1.3\textwidth, font={scriptsize}} %调整表格标题宽度
\caption{\textbf{NMI of clustering methods in different scenarios.} For each comparison, the best result is highlighted in boldface.}
%\scriptsize
%\hspace{-20mm}
\scalebox{0.65}{%放缩
\renewcommand{\arraystretch}{2} %调行间距
\tabcolsep 1mm %space between two columns. 用于调整列间距
\begin{tabular}{lccccccccc}
\hline\hline

\multirow{2}{*}{Dataset} &\multirow{2}{*}{$p$} &\multicolumn{2}{c}{Centralized setting} &\multicolumn{6}{c}{Federated setting}\\ \cmidrule(r){3-4} \cmidrule(r){5-10}
\quad &\quad &\textcolor{mygray}{KM} &\textcolor{mygray}{FCM} &k-FED &FFCM &SDA-FC-KM &SDA-FC-FCM &PPFC-GAN &CCFC(ours)\\

\hline
\multirow{5}{*}{MNIST} &0.0 &\multirow{5}{*}{\textcolor{mygray}{0.5304}} &\multirow{5}{*}{\textcolor{mygray}{0.5187}} &0.5081 &0.5157 &0.5133 &0.5141 &0.6582 &\textbf{0.9236}\\
\quad &0.25 &\quad &\quad &0.4879 &0.5264 &0.5033 &0.5063 &0.6392 &\textbf{0.8152} \\
\quad &0.5 &\quad &\quad &0.4515 &0.4693 &0.5118 &0.5055 &\textbf{0.6721} &0.6718 \\
\quad &0.75 &\quad &\quad &0.4552 &0.4855 &0.5196 &0.5143 &\textbf{0.7433} &0.3611 \\
\quad &1.0 &\quad &\quad &0.4142 &0.5372 &0.5273 &0.5140 &\textbf{0.8353} &0.0766 \\

\hline
\multirow{5}{*}{Fashion-MNIST} &0.0 &\multirow{5}{*}{\textcolor{mygray}{0.6070}} &\multirow{5}{*}{\textcolor{mygray}{0.6026}} &0.5932 &0.5786 &0.5947 &0.6027 &0.6091 & \textbf{0.6237} \\
\quad &0.25 &\quad &\quad &0.5730 &0.5995 &\textbf{0.6052} &0.5664 &0.5975 & 0.5709\\
\quad &0.5 &\quad &\quad &0.6143 &\textbf{0.6173} &0.6063 &0.6022 &0.5784 & 0.6023\\
\quad &0.75 &\quad &\quad &0.5237 &\textbf{0.6139} &0.6077 &0.5791 &0.6103 &0.4856\\
\quad &1.0 &\quad &\quad &0.5452 &0.5855 &0.6065 &0.6026 &\textbf{0.6467} &0.1211\\

\hline
\multirow{5}{*}{CIFAR-10} &0.0 &\multirow{5}{*}{\textcolor{mygray}{0.0871}} &\multirow{5}{*}{\textcolor{mygray}{0.0823}} &0.0820 &0.0812 &0.0823 &0.0819 &0.1165 &\textbf{0.2449}\\
\quad &0.25 &\quad &\quad &0.0866 &0.0832 &0.0835 &0.0818 &0.1185 &\textbf{0.2094} \\
\quad &0.5 &\quad &\quad &0.0885 &0.0870 &0.0838 &0.0810 &0.1237 &\textbf{0.2085}\\
\quad &0.75 &\quad &\quad &0.0818 &0.0842 &0.0864 &0.0808 &0.1157 &\textbf{0.1189}\\
\quad &1.0 &\quad &\quad &0.0881 &0.0832 &0.0856 &0.0858 &\textbf{0.1318} &0.0639\\

\hline
\multirow{5}{*}{STL-10} &0.0 &\multirow{5}{*}{\textcolor{mygray}{0.1532}} &\multirow{5}{*}{\textcolor{mygray}{0.1469}} &0.1468 &0.1436 &0.1470 &0.1406 &0.1318 &\textbf{0.2952}\\
\quad &0.25 &\quad &\quad &0.1472 &0.1493 &0.1511 &0.1435 &0.1501 &\textbf{0.1727}\\
\quad &0.5  &\quad &\quad &0.1495 &0.1334 &0.1498 &0.1424 &0.1432 &\textbf{0.2125}\\
\quad &0.75 &\quad &\quad &0.1455 &0.1304 &0.1441 &0.1425 &0.1590 &\textbf{0.1610}\\
\quad &1.0  &\quad &\quad &0.1403 &0.1565 &0.1477 &0.1447 &\textbf{0.1629} &0.0711\\

\hline
count &- &- &- &0 &2 &1 &0  &6 &11\\
\hline\hline
\end{tabular}\label{NMI}}
\end{table*}

\begin{table*}[!t]
%\vspace{-3cm}
%\setlength{\abovecaptionskip}{10pt}%
%\setlength{\belowcaptionskip}{10pt}%
%\captionsetup{width=1.3\textwidth, font={scriptsize}} %调整表格标题宽度
\caption{\textbf{Kappa of clustering methods in different scenarios.} For each comparison, the best result is highlighted in boldface.}
%\scriptsize
%\hspace{-20mm}
\scalebox{0.65}{%放缩
\renewcommand{\arraystretch}{2} %调行间距
\tabcolsep 1mm %space between two columns. 用于调整列间距
\begin{tabular}{ccccccccccc}
\hline\hline

\multirow{2}{*}{Dataset} &\multirow{2}{*}{$p$} &\multicolumn{2}{c}{Centralized setting} &\multicolumn{6}{c}{Federated setting}\\ \cmidrule(r){3-4} \cmidrule(r){5-10}
\quad &\quad &\textcolor{mygray}{KM} &\textcolor{mygray}{FCM} &k-FED &FFCM &SDA-FC-KM &SDA-FC-FCM &PPFC-GAN &CCFC(ours)\\

\hline
\multirow{5}{*}{MNIST} &0.0 &\multirow{5}{*}{\textcolor{mygray}{0.4786}} &\multirow{5}{*}{\textcolor{mygray}{0.5024}} &0.5026 &0.5060 &0.4977 &0.5109 &0.6134 &\textbf{0.9619}\\
\quad &0.25 &\quad &\quad &0.4000 &0.5105 &0.4781 &0.5027 &0.5773 &\textbf{0.8307} \\
\quad &0.5 &\quad &\quad &0.3636 &0.3972 &0.4884 &0.4967 &0.6007 &\textbf{0.6534} \\
\quad &0.75 &\quad &\quad &0.3558 &0.4543 &0.4926 &0.5021 &\textbf{0.6892} &0.3307 \\
\quad &1.0 &\quad &\quad &0.3386 &0.5103 &0.5000 &0.5060 &\textbf{0.7884} &0.0911 \\

\hline
\multirow{5}{*}{Fashion-MNIST} &0.0 &\multirow{5}{*}{\textcolor{mygray}{0.4778}} &\multirow{5}{*}{\textcolor{mygray}{0.5212}} &0.4657 &0.4974 &0.4640 &0.5252 &0.4857 & \textbf{0.6411}\\
\quad &0.25 &\quad &\quad &0.5222 &0.5180 &0.4763 &0.4962 &0.4721 &\textbf{0.5261}\\
\quad &0.5 &\quad &\quad &0.4951 &0.4974 &0.4826 &0.5258 &0.4552 &\textbf{0.5929}\\
\quad &0.75 &\quad &\quad &0.4240 &\textbf{0.4995} &0.4774 &0.4955 &0.4774 &0.3945\\
\quad &1.0 &\quad &\quad &0.3923 &0.4672 &0.4825 &0.5323 &\textbf{0.5745} &0.1434\\

\hline
\multirow{5}{*}{CIFAR-10} &0.0 &\multirow{5}{*}{\textcolor{mygray}{0.1347}} &\multirow{5}{*}{\textcolor{mygray}{0.1437}} &0.1305 &0.1439 &0.1275 &0.1283 &0.1426 &\textbf{0.2854}\\
\quad &0.25 &\quad &\quad &0.1366 &0.1491 &0.1275 &0.1376 &0.1400 &\textbf{0.2281}\\
\quad &0.5 &\quad &\quad &0.1252 &0.1316 &0.1307 &0.1411 &0.1443 &\textbf{0.2214}\\
\quad &0.75 &\quad &\quad &0.1303 &0.1197 &0.1360 &\textbf{0.1464} &0.1358 &0.1214\\
\quad &1.0 &\quad &\quad &0.1147 &0.1237 &0.1341 &0.1494 &\textbf{0.1499} &0.1047\\

\hline
\multirow{5}{*}{STL-10} &0.0 &\multirow{5}{*}{\textcolor{mygray}{0.1550}} &\multirow{5}{*}{\textcolor{mygray}{0.1602}} &0.1390 &0.1514 &0.1533 &0.1505 &0.1557 &\textbf{0.1687}\\
\quad &0.25 &\quad &\quad &0.1361 &0.1479 &0.1448 &0.1527 &\textbf{0.1611} &0.1422\\
\quad &0.5 &\quad &\quad &0.1505 &0.1112 &0.1377 &\textbf{0.1620} &0.1415  &0.1407\\
\quad &0.75 &\quad &\quad &0.1256 &0.1001 &0.1513 &0.1603 &\textbf{0.1813} &0.1133\\
\quad &1.0 &\quad &\quad &0.1328 &0.1351 &0.1527 &0.1553 &\textbf{0.1868}  &0.0519\\

\hline
count &- &- &- &0 &1 &0  &2  &7 &10\\
\hline\hline
\end{tabular}\label{Kappa}}
\end{table*}

\subsection{Clustering Performance Comparison}
Federated clustering, as an emerging field, currently lacks a uniform evaluation framework. For example, none of the benchmark datasets used in studies \cite{dennis2021heterogeneity,stallmann2022towards,wang2022federated} are consistent, making direct comparisons difficult. Thus, considering the availability of baseline codes and ease of implementation, we select k-FED \cite{dennis2021heterogeneity}, FFCM \cite{stallmann2022towards}, SDA-FC-KM \cite{sdafc}, SDA-FC-FCM \cite{sdafc} and PPFC-GAN \cite{ppfcgan} as baselines to validate the effectiveness of CCFC. To avoid excessive hyperparameter tuning, we use specific hyperparameter settings for each method across various simulated scenarios within the same dataset.

The clustering performance is assessed based on two metrics, NMI \cite{strehl2002cluster} and Kappa \cite{liu2019evaluation}, as presented in \cref{NMI} and \cref{Kappa}, respectively. One can see that: 1) Although the two metrics exhibit different ranks in some federated scenarios, the advantages of the proposed method are quite significant in most scenarios. In some scenarios, its performance scores even double those of the best baseline methods. The most conspicuous case is observed on MNIST with the data heterogeneity $p = 0$, where it achieves an NMI score surpassing PPFC-GAN by 0.2654 and a Kappa score exceeding 0.3485. 2) CCFC suffers from data heterogeneity, while the synthetic data-aided methods (SDA-FC-KM, SDA-FC-FCM, and PPFC-GAN) are immune to, and can even benefit from, data heterogeneity. Thus, performing contrastive clustering on shared synthetic data appears to offer a solution to mitigate the data heterogeneity problem without sharing private data. However, the increased utility of synthetic data often comes with a heightened risk of deep data leakage from shared synthetic data. In contrast to synthetic data, which may serve as replicas of real data and thus pose a direct privacy risk \cite{ghalebikesabi2023differentially}, sharing model parameters offers a significantly greater degree of privacy preservation. Recall that the core distinction between FC and centralized clustering lies in privacy preservation. Hence, we choose to use the FedAvg framework \cite{mcmahan2017communication} to perform contrastive clustering, rather than SDA-FC \cite{sdafc}. As widely acknowledged, the issue of data heterogeneity remains unresolved and challenging \cite{shao2023survey, ma2022state, lu2024federated}. Thus, this work will emphasize how contrastive learning aids in FC, while the issue of data heterogeneity is left for future research. 3) Representation learning is poised to become the primary catalyst for progress in federated clustering. The most related method to our proposed CCFC is k-fed, with the main distinction being the inclusion of a representation learning process in CCFC. Surprisingly, this learning process results in substantial NMI score improvements of up to 0.4155 and Kappa score enhancements of 0.4593 on the most conspicuous case. In the next subsection, we will provide a deeper analysis to illustrate how representation learning affects federated clustering tasks.

%Exploring how to balance this trade-off could be an intriguing avenue for future research.

%an issue that remains unresolved and challenging \cite{shao2023survey, bai2021training, somepalli2023diffusion, meng2024diffusion, zhang2024generate}
%the issue of deep data leakage from shared synthetic data remains unresolved and challenging

%Despite being more sensitive to data heterogeneity than synthetic data-aided methods (SDA-FC-KM, SDA-FC-FCM, and PPFC-GAN), CCFC shows overwhelming superiority in low data heterogeneity scenarios.

%Although the synthetic data can make the model immune to the data heterogeneity problem without sharing private data,

%The proposed method CCFC is sensitive to the data heterogeneity problem, whereas synthetic data aided methods (SDA-FC-KM, SDA-FC-FCM and PPFC-GAN) are not only immune to such a problem, but may even benefit from increased data heterogeneity. Hence, training the model on shared synthetic data appears to offer a solution to the data heterogeneity problem encountered by CCFC. However, the enhanced utility of synthetic data is often accompanied by an elevated risk of privacy leakage \cite{shao2023survey, bai2021training}. Exploring how to balance this trade-off could be an intriguing avenue for future research. 3) The proposed method demonstrates a higher potential.

%1. 表现最好
%2. 联邦表示学习有望推动联邦聚类的进步
%3. 尽管PPFC-GAN稳定, 但天花板低。 non-iid问题

\begin{figure}[!t]
\centering
%\vspace{-3cm}
\subfigure[Raw pixels]{
\includegraphics[height = 4cm, width = 6cm]{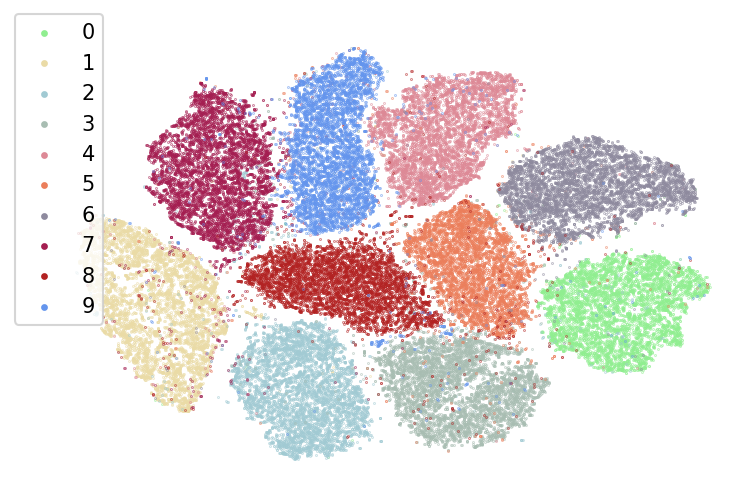}}
\quad
\subfigure[PPFC-GAN]{
\includegraphics[height = 4cm, width = 6cm]{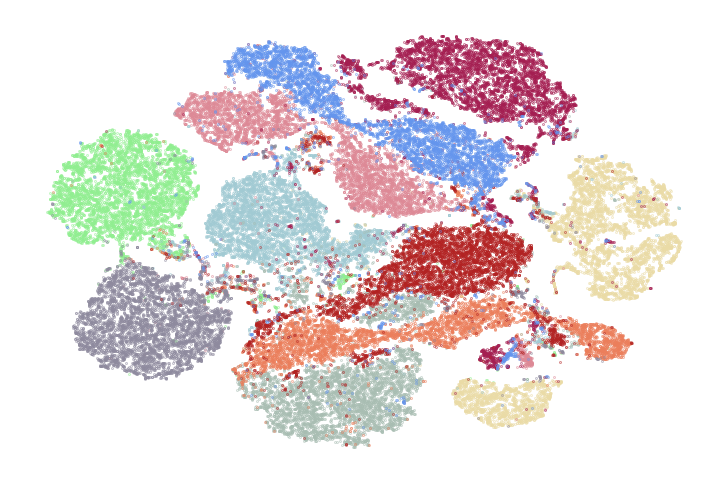}}

\subfigure[SCFC]{
\includegraphics[height = 4cm, width = 6cm]{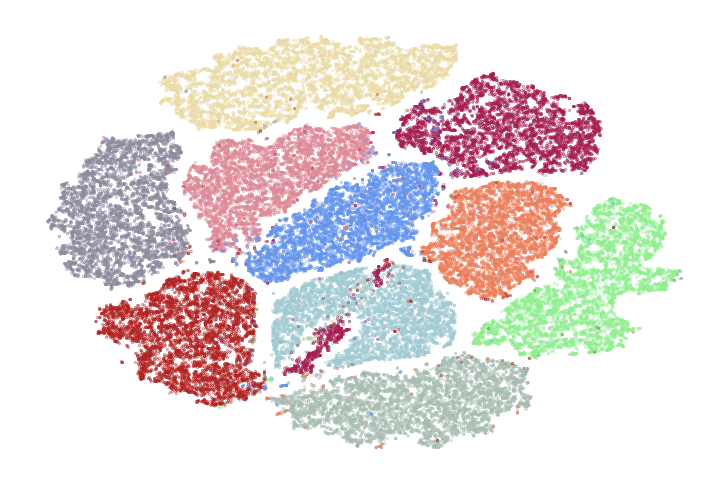}}
\quad
\subfigure[CCFC]{
\includegraphics[height = 4cm, width = 6cm]{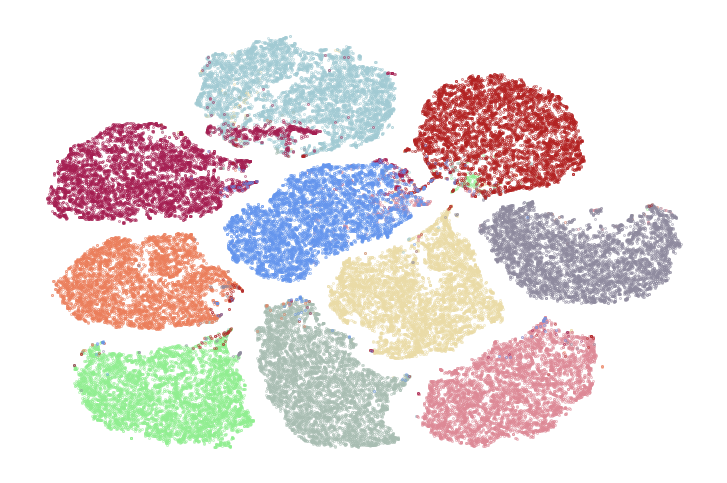}}
\caption{t-SNE visualizations on MNIST with the data heterogeneity $p = 0$ (best viewed in color and upon zooming in).}
\label{case2}
\end{figure}

\subsection{Effectiveness Analysis of Representation Learning}

In centralized clustering, the strength of the representation learning process is its capacity to map raw data features into more clustering-friendly latent representations that exhibit improved intra-cluster compactness and inter-cluster separability \cite{zhou2022comprehensive, ren2024deep}. In FC, we will maintain this belief and validate it by qualitatively and quantitatively analyzing the clustering-friendliness of the learned representations in different federated scenarios.

\begin{table}[!t]
\centering
%\vspace{-3cm}
%\setlength{\abovecaptionskip}{0pt}%
%\setlength{\belowcaptionskip}{10pt}%
\caption{MNIST test accuracy (\%) for kNN on different representation spaces.}
%\hspace{-28mm}
%\scalebox{0.89}{%放缩
\renewcommand{\arraystretch}{1.5} %行间距
\tabcolsep 1.9mm %列间距
\begin{tabular}{lccccccc}
\hline\hline
k  &3 &5 &7 &9 &100 &1000 &7000  \\\hline
Raw pixels &97.05 &96.88 &96.94 &96.59 & 94.40 & 87.31 &70.16                      \\
PPFC-GAN & 95.26 & 95.20 & 95.25 & 95.25  & 92.73 & 85.37 &71.19                          \\
SCFC & 98.11 & 98.23 & 98.25 & 98.21 & 97.54 & 96.30 &94.09                    \\
CCFC & \textbf{98.25} & \textbf{98.33} & \textbf{98.26} & \textbf{98.32} & \textbf{97.86} & \textbf{97.09}  &\textbf{97.09}                          \\
\hline\hline
\end{tabular}%}
\label{case3}
\end{table}

\subsubsection{IID scenario}
In this context, we explore both qualitatively and quantitatively the latent representations learned by PPFC-GAN and CCFC on the MNIST dataset with data heterogeneity $p = 0$. Moreover, to demonstrate the efficacy of the cluster-contrastive learning, we construct an additional baseline method, named sample-contrastive federated clustering (SCFC), achieved by reverting the cluster-contrastive model to SimSiam. This signifies that, throughout the local training process, SCFC abstains from leveraging the semantic structures discovered by clustering.

\noindent \textbf{Qualitative analysis.}
As shown in \cref{case2}, one can observer that: 1) In the raw data space, there are 10 discernible modes, yet many samples from different categories are mixed up together, necessitating representation learning. 2) Representation learning through contrastive learning significantly alleviates the mixing problem, and the cluster-contrastive learning yields more compact representations. 3) Interestingly, the latent representations learned by PPFC-GAN seem to exacerbate the mixing problem, implying that the advantages of PPFC-GAN may not be attributed to the acquisition of more clustering-friendly representations. Hence, to thoroughly figure out this matter, further quantitative analysis is necessary.

\noindent \textbf{Quantitative analysis.}
To offer more compelling evidence that representation learning can mitigate the mixing problem in the original data space, we run the k-nearest neighbors (KNN) classifier in different representation spaces. The better the test performance, the fewer samples from different categories among the $k$ neighbors, suggesting a reduced severity of the mixing problem.

As shown in \cref{case3}, one can find that: 1) The test performance in the latent spaces is comparable to, or in some cases, inferior to that in the original data space for small values of $k$. However, as $k$ values increase, the merits of latent representation learning progressively come to fruition by mitigating the mixing problem. 2) The proposed method demonstrates superior effectiveness and robustness, better preserving the inherent semantic structure of the data.

\begin{figure*}[!t]
\centering
%\vspace{-3cm}
\subfigure{
\includegraphics[height = 4cm, width = 4cm]{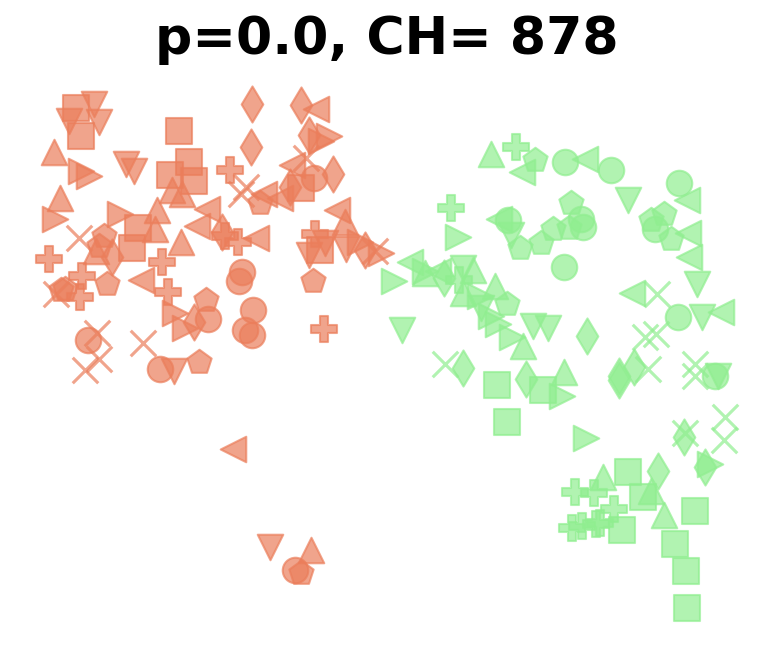}
}
\,
\subfigure{
\includegraphics[height = 4cm, width = 4cm]{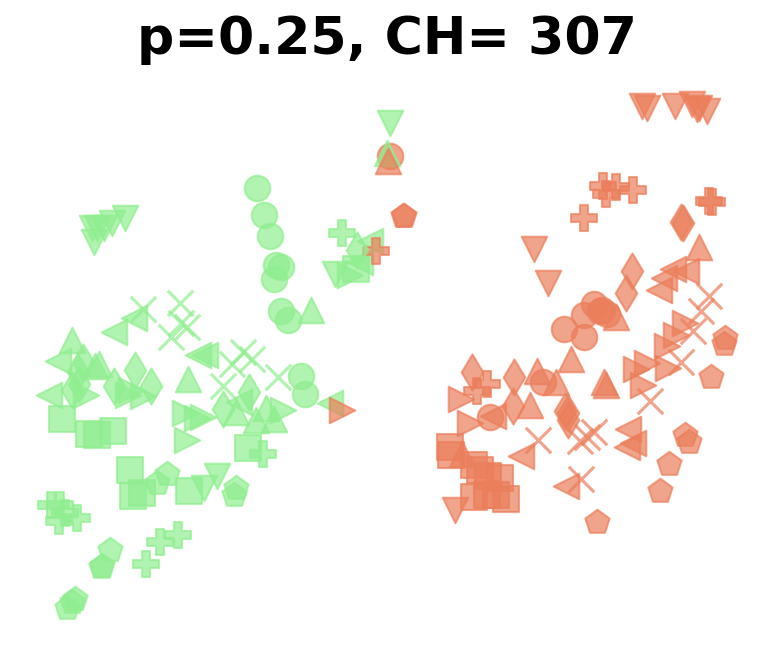}
}
\,
\subfigure{
\includegraphics[height = 4cm, width = 4cm]{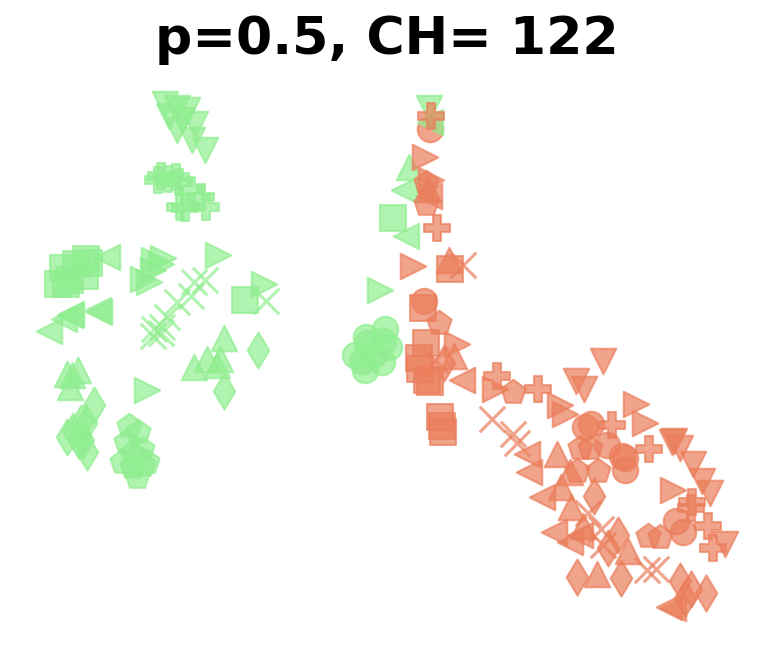}
}

\subfigure{
\includegraphics[height = 4cm, width = 4cm]{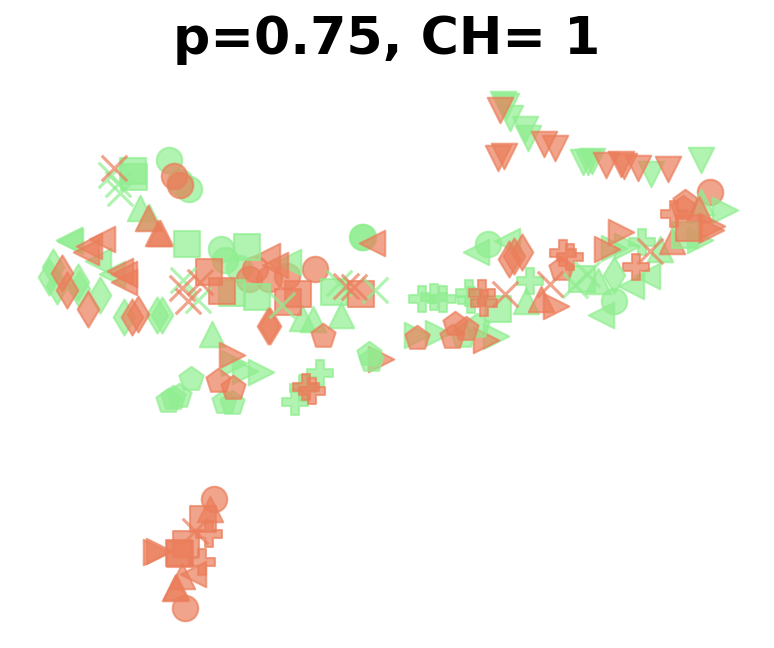}
}
\quad
\subfigure{
\includegraphics[height = 4.1cm, width = 6cm]{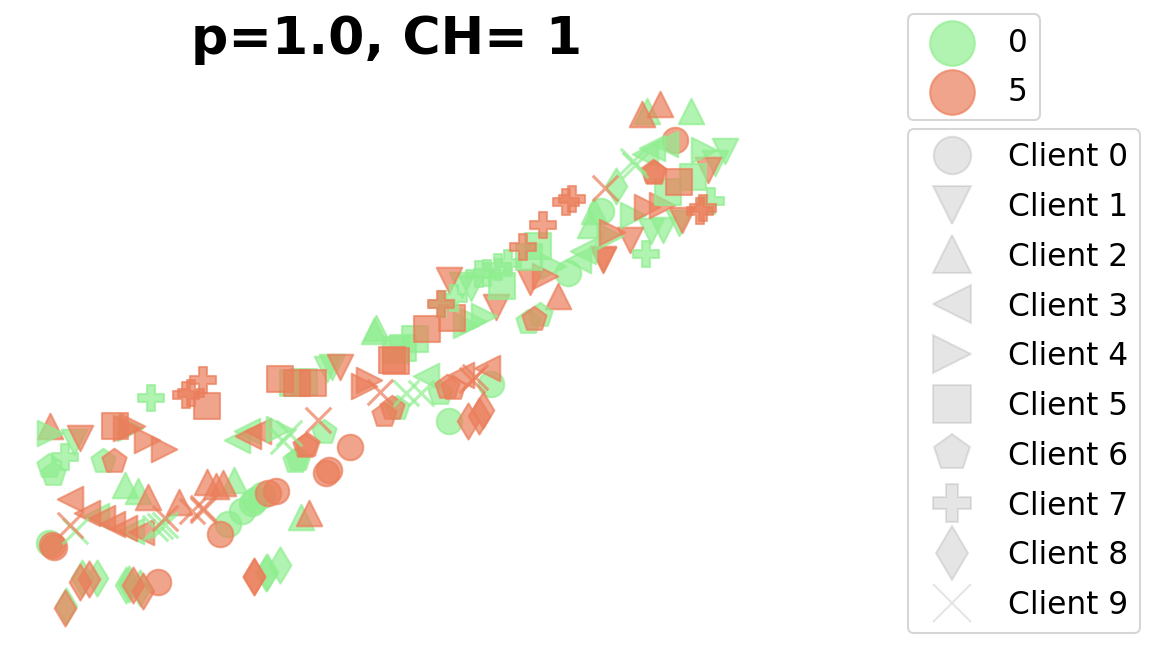}
}
\caption{\textbf{t-SNE visualization of learned representations under data heterogeneity in MNIST (best viewed in color)}. Each color corresponds to a specific category of digits in MNIST,  and each grey marker shape denotes a specific local model trained on private data from the corresponding client. A greater $p$ implies increased data heterogeneity, while a higher CH score denotes more clustering-friendly representations that show improved intra-cluster compactness and inter-cluster separability. However, as $p$ increases, the learned representations gradually lose their clustering-friendly traits.}
\label{case00}
\end{figure*}

%证明了表示学习在最简单的FC场景中的有效性，大多数场景都是非独立同分布的，实践意义重大。
\subsubsection{Non-IID scenarios}
While the effectiveness of the representation learning process in CCFC has been confirmed in the simplest federated scenario (IID scenario), and resolving data heterogeneity in FC is not the focus of this work, it is essential to analyze how data heterogeneity impairs this process for the future enhancement of the method.

To this end, we first randomly select two classes of digits in MNIST, each containing 100 samples, distributed across 10 clients with varying data heterogeneity $p$. Then, we visualize the latent representations learned by the local models of CCFC, and report their Calinski-Harabasz (CH) scores \cite{calinski1974dendrite} based on the true labels to measure the corresponding clustering-friendly properties. A higher CH score denotes more clustering-friendly representations that show improved intra-cluster compactness and inter-cluster separability. As shown in \cref{case00}, one can see that: 1) The learned representations gradually lose their clustering-friendly properties with an increase in data heterogeneity, indicated by decreased CH scores. 2) Heterogeneous data leads to \textit{semantic inconsistency} in the latent spaces of local models. In the IID scenario ($p = 0$), the latent representations of the same digit, learned by different local models, are intermingled, with their spatial distribution unaffected by the specific client. In contrast, under non-IID conditions (e.g., $p = 0.5$), the spatial distribution of the learned representations for the same digit exhibits pronounced client-specific characteristics. 3) Heterogeneous data weakens the discriminative ability of contrastive learning. With the increase in data heterogeneity $p$, the expansion of the major class within local data reduces the available discriminative information during comparison, hindering the differentiation between categories and resulting in overlapping cluster representations, i.e., \textit{cluster collapse}.

The above discussion explains the reasons behind the overwhelming advantage of the proposed method in the IID scenario and how data heterogeneity undermines its effectiveness. Addressing the challenges of semantic inconsistency and cluster collapse induced by data heterogeneity may further unlock the method's potential, which is left for future work.
%判别性消失

\begin{table*}[!t]
\centering
%\vspace{-3cm}
%\setlength{\abovecaptionskip}{10pt}%
%\setlength{\belowcaptionskip}{10pt}%
%\captionsetup{width=1.3\textwidth, font={scriptsize}} %调整表格标题宽度
\caption{\textbf{Kappa of each client in one experiment.} Imbalance ratio (IR) denotes the ration of the sample size of the largest class to the smallest class.}
%\scriptsize
%\hspace{-20mm}
\scalebox{0.6}{%放缩
\renewcommand{\arraystretch}{2} %调行间距
\tabcolsep 4.5mm %space between two columns. 用于调整列间距
\begin{tabular}{cccccccc}
\hline\hline

\multirow{2}{*}{Client id} &\multirow{2}{*}{IR} &\multicolumn{2}{c}{Standalone models} &\multicolumn{2}{c}{FedAvg models} &\multicolumn{2}{c}{FedAvg models with model-contrastive learning}\\ \cmidrule(r){3-4} \cmidrule(r){5-6} \cmidrule(r){7-8}
\quad &\quad &SCFC$^{\ddag}$ &CCFC$^{\ddag}$ &SCFC$^{\dag}$ &CCFC$^{\dag}$ &SCFC &CCFC\\

\hline
1  &1.18   &0.1286 & \textbf{0.4255} & 0.8243 & \textbf{0.8749} & 0.8970 & \textbf{0.9605}\\
2  &1.23   & 0.1774 & \textbf{0.2036} & 0.8297 & \textbf{0.8787} & 0.9046 & \textbf{0.9643}\\
3  &1.39   & \textbf{0.1685} & 0.0517 & 0.8273 & \textbf{0.8783} & 0.9027 & \textbf{0.9611} \\
4  &1.21   & \textbf{0.2751} & 0.1847 & 0.8239 & \textbf{0.8714} & 0.8987 & \textbf{0.9617}\\
5  &1.24   & \textbf{0.2134} & 0.0926 & 0.8179 & \textbf{0.8689} & 0.8976 & \textbf{0.9609}\\
6  &1.23   & 0.1862 & \textbf{0.1892} & 0.8201 & \textbf{0.8696} & 0.8987 & \textbf{0.9625}\\
7  &1.19   & \textbf{0.1876} & 0.1681 & 0.8241 & \textbf{0.8722} & 0.8995 & \textbf{0.9624}\\
8  &1.19   & 0.3407 & \textbf{0.3741} & 0.8330 & \textbf{0.8750} & 0.9030 & \textbf{0.9611} \\
9  &1.18   & 0.2810 & \textbf{0.2936} & 0.8190 & \textbf{0.8687} & 0.8983 & \textbf{0.9616}\\
10 &1.53   & 0.1940 & \textbf{0.2727} & 0.8214 & \textbf{0.8686} & 0.9000 & \textbf{0.9630} \\
\hline\hline
\end{tabular}}
\label{case4}
\end{table*}

\subsection{Ablation Study}
Recall that CCFC comprises three key components: \textit{cluster-contrastive learning, FedAvg and the model-contrastive regularization term}. To validate the effectiveness of each component, we devise four additional baselines within two distinct federated clustering frameworks:

\noindent \textbf{1) Standalone models.}  In this set of baselines, each client independently performs sample-contrastive learning (denoted as SCFC$^{\ddag}$) or cluster-contrastive learning (denoted as CCFC$^{\ddag}$).

\noindent \textbf{2) FedAvg models.} In this set of baselines, each client interactively performs sample-contrastive learning (denoted as SCFC$^{\dag}$) or cluster-contrastive learning (denoted as CCFC$^{\dag}$) within the FedAvg framework, without employing the model-contrastive regularization term.

\cref{case4} shows that: 1) The complete model exhibits the most favorable clustering performance, implying that the removal of any component can result in performance degradation. 2) The most substantial improvements arise from the collaborative training among clients, while cluster-contrastive learning and the model-contrastive regularization term can yield further performance improvements.

\begin{figure*}
\centering
\includegraphics[height = 6cm, width = 13cm]{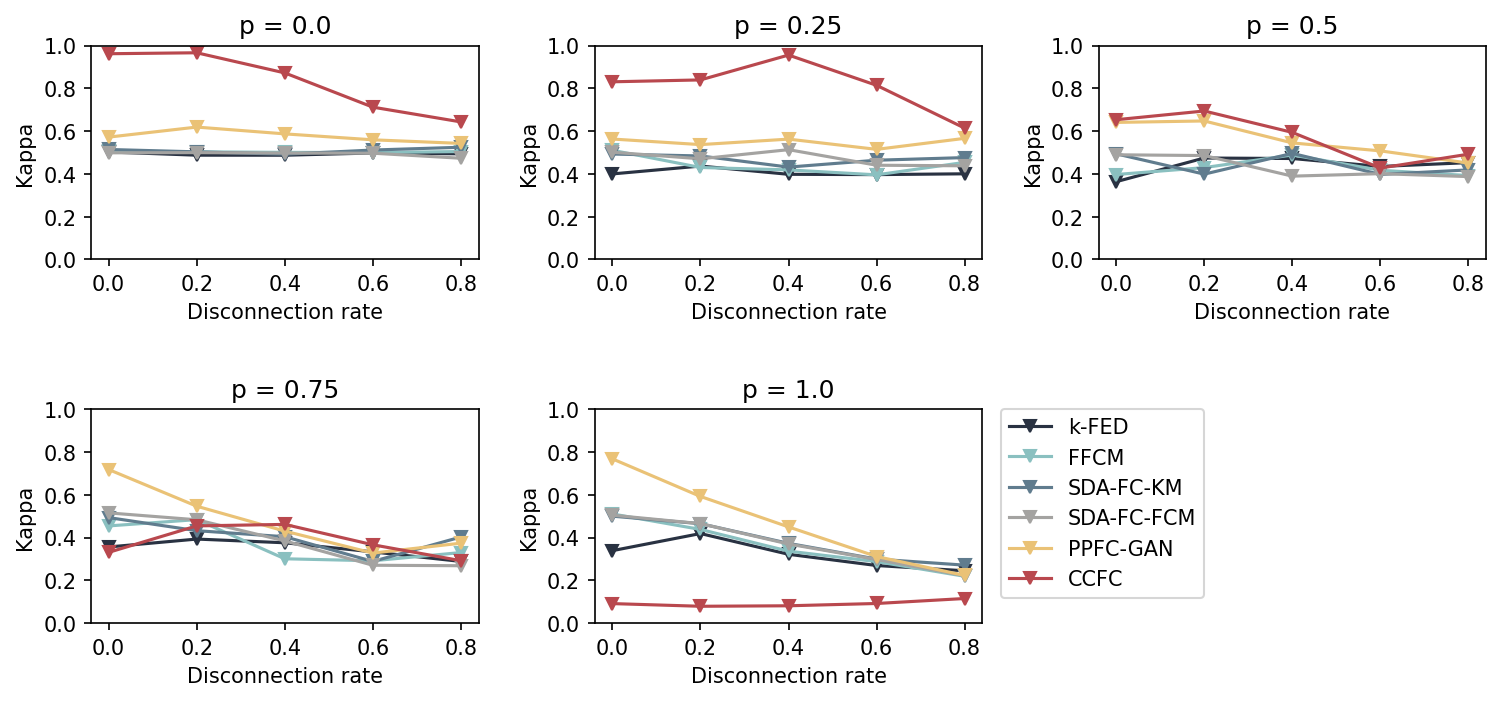}
\caption{Sensitivity of CCFC to the device failures on MNIST with different data heterogeneity.}\label{case6}
\end{figure*}

\begin{figure}[!t]
\centering
%\vspace{-3cm}
\subfigure[NMI]{
\includegraphics[height = 3.5cm, width = 4cm]{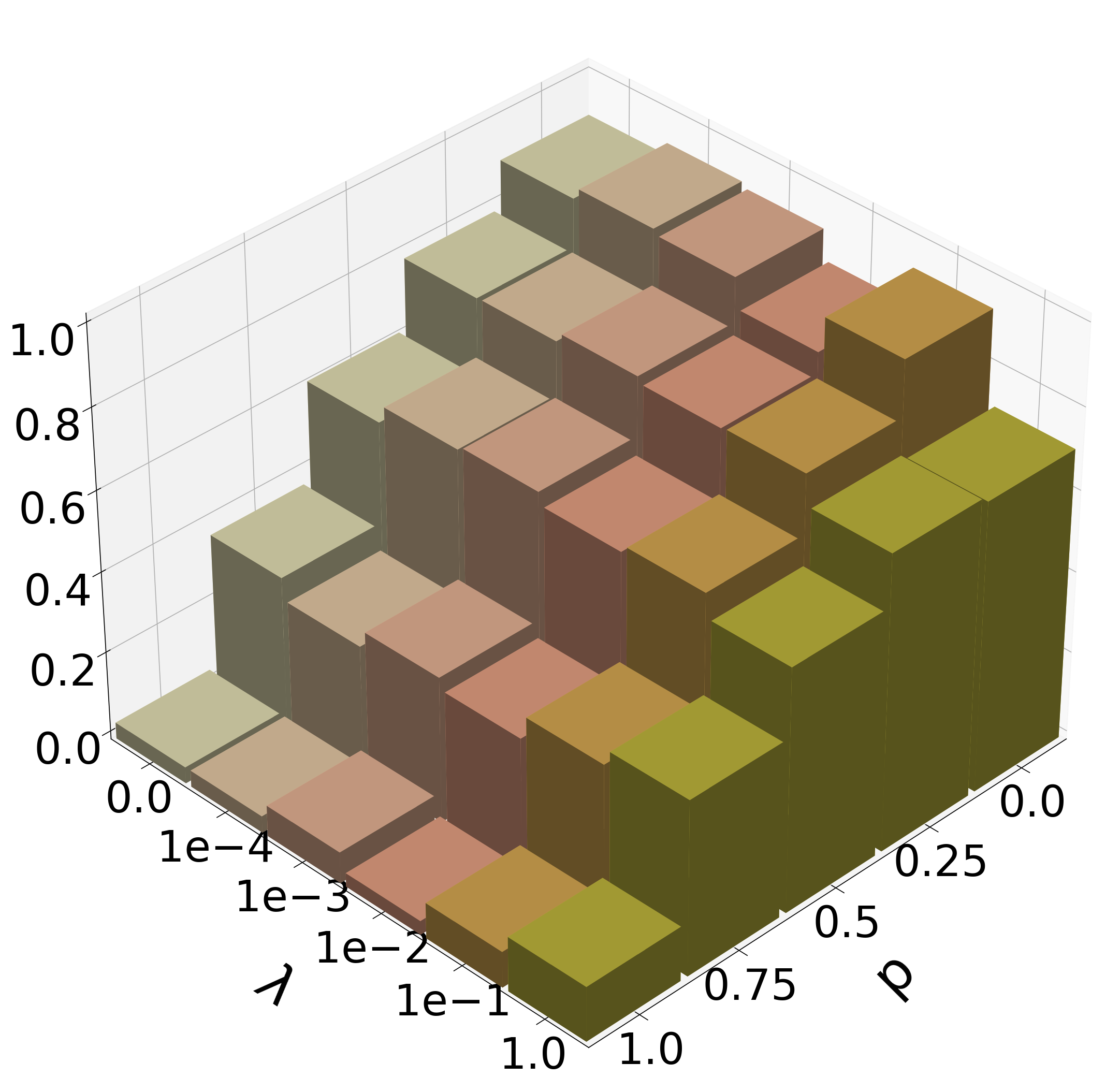}}
\quad
\subfigure[Kappa]{
\includegraphics[height = 3.5cm, width = 4cm]{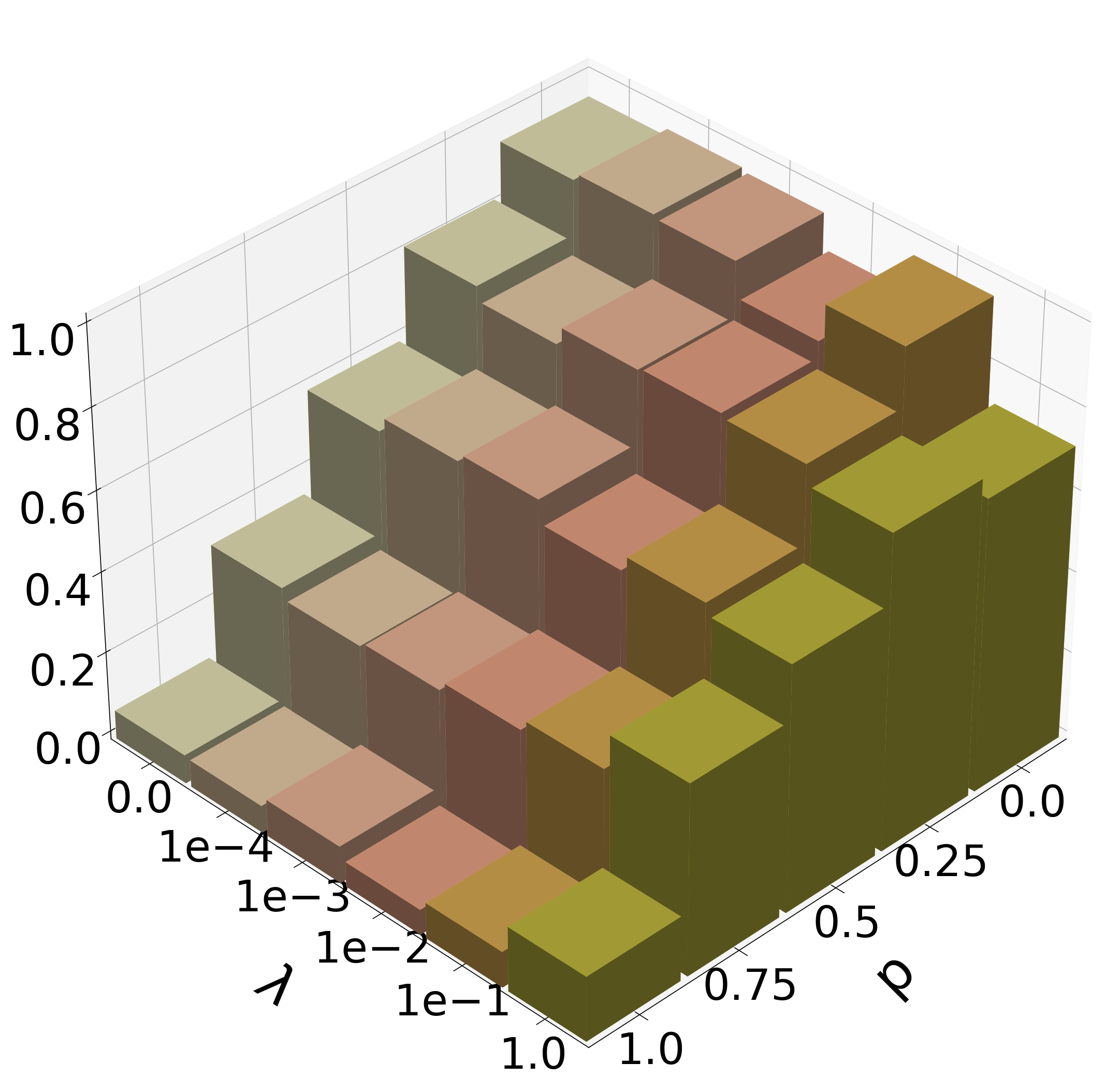}}
\caption{Sensitivity of CCFC to $\lambda$ on MNIST with different data heterogeneity.}
\label{case5}
\end{figure}

\subsection{Hyperparameter Sensitivity Analysis}
Although we have previously validated the efficacy of the model-contrastive regularization term, the relative importance of this term compared to the cluster-contrastive loss remains unexplored. To this end, we conduct a series of CCFC experiments with different $\lambda$ on MNIST.

\cref{case5} reveals that: 1) The clustering performance remains robust across a wide range of $\lambda$ under a fixed federated heterogeneous scenario. 2) The optimal value of $\lambda$ differs across different federated scenarios. Consequently, the performance of CCFC within \cref{NMI} and \cref{Kappa} is somewhat underestimated, as we used the same $\lambda$ for different federated scenarios of the same dataset to prevent excessive hpyerparameter tuning. 3) The sensitivity of the model to data heterogeneity is more pronounced compared to $\lambda$.

%大范围内鲁棒
%不同异质性设定下的最优lambda是不同的。

\subsection{Device Failures}
In real-world scenarios, client devices may occasionally fail to connect with the server during the training process due to wireless network fluctuations, energy constraints, etc. As a result, specific data characteristics from the failed devices may be lost, resulting in poor and unrobust performance.  Hence, it is essential to investigate the sensitivity of CCFC to device failures  from a practical viewpoint.

Following \cite{sdafc, ppfcgan}, we simulate different disconnection scenarios by scaling the \textbf{disconnection rate}, which measures the proportion of disconnected clients among all clients. Throughout the training process, only the connected clients are involved. As shown in \cref{case6}, while CCFC exhibits greater sensitivity to device failures compared to the baselines, it remains the best performance in most cases.
%表现最好
%不仅受益于异质数据量的增加,受益于同质数据量的增加,潜力更大

In summary: 1) The representations acquired through the proposed cluster-contrastive learning model are more clustering-friendly. 2) Benefiting from the more clustering-friendly representations, CCFC demonstrates superior performance. 3) With the rise of data heterogeneity, the learned representations progressively lose their clustering-friendly traits, as evidenced by semantic inconsistency in the latent spaces of local models and overlapping cluster representations (cluster collapse). 4) The considered key components within CCFC are all essential for its superiority. 5) The clustering performance remains robust across a wide range of the tradeoff hyperparameter $\lambda$. 6) While CCFC exhibits greater sensitivity to device failures compared to the baselines, it remains the best performance in most cases.

%所提出的簇对比学习模型学出来的表示更加聚类友好
%受益于表示学习，所提方法表现好。
%3个元素必不可少
%敏感性分析
%device failures

\section{Conclusion}
\label{sect5}
As widely acknowledged, the performance of federated methods can be hindered by the issue of data heterogeneity, and this problem remains unresolved and challenging \mbox{\cite{shao2023survey, ma2022state, lu2024federated}.} Thus, this work focuses on how contrastive learning aids in FC and attempts to analyze how data heterogeneity impairs the clustering-friendliness of the learned representations.

To this end, we first propose a cluster-contrastive model for learning more clustering-friendly representations, and introduce a simple yet effective federated clustering method by embedding the cluster-contrastive model into the FedAvg framework. Comprehensive experiments demonstrate the superiority of the cluster-contrastive model in learning clustering-friendly representations, as well as the excellence of the proposed clustering method in terms of clustering performance and handling device failures. The most related method to our CCFC is k-fed \cite{dennis2021heterogeneity}, with the main distinction being the inclusion of a representation learning process in CCFC. Surprisingly, this learning process results in substantial NMI \cite{strehl2002cluster} score improvements of up to 0.4155 and Kappa \cite{liu2019evaluation} score enhancements of 0.4593 on the most conspicuous case. Therefore, representation learning is likely to emerge as the key driver of future advancements in FC.

Then, we qualitatively and quantitatively analyze the clustering-friendliness of the learned representations in different federated scenarios. With the rise of data heterogeneity, the learned representations progressively lose their clustering-friendly traits, as evidenced by \textit{semantic inconsistency} in the latent spaces of local models and overlapping cluster representations (\textit{cluster collapse}). Resolving these two unfavorable phenomena may further unleash the potential of CCFC, which is left for future work.

We hope our work will draw the community's focus to the fundamental role of representation learning in FC, and lay a foundation for future advancements within the community.

\bibliographystyle{elsarticle-num}
\bibliography{references.bib}

\end{document}